\documentclass[10pt,twocolumn,letterpaper]{article}

\usepackage{iccv}
\usepackage{times}
\usepackage{epsfig}
\usepackage{graphicx}
\usepackage{amsmath}
\usepackage{amssymb}

\usepackage[breaklinks=true,bookmarks=false]{hyperref}

\iccvfinalcopy 

\def\iccvPaperID{****} 

\begin{document}

\title{A Jointly Learned Deep Architecture for Facial Attribute Analysis and Face Detection in the Wild}

\author{Keke He, Yanwei Fu, Xiangyang Xue\\
Fudan University \\
{\tt\small \{kkhe15, yanweifu, xyxue\}@fudan.edu.cn}
}

\maketitle

\begin{abstract}
Facial attribute analysis in the real world scenario is very challenging mainly because of complex face variations. Existing works of analyzing face attributes are mostly based on the cropped and aligned face images. However, this result in the capability of attribute prediction heavily relies on the preprocessing of face detector. To address this problem, we present a novel jointly learned deep architecture for both facial attribute analysis and face detection. Our framework can process the natural images in the wild and our experiments on CelebA and LFWA datasets clearly show that the state-of-the-art performance is obtained. 
\end{abstract}

\section{Introduction}
The problem of automatically analyzing the facial attributes received increasing attention recently
\cite{rudd2016moon,liu2015deep,off_shelf_face}.
Face attributes may potentially benefit a number of real-world applications,
such as face alignment\cite{ren2014face,sagonas2013300,zhang2014facial,zhou2013extensive},
head pose estimation \cite{zhu2012face,yang2015face} and face verification
\cite{taigman2014deepface,sun2014deep}. Analyzing facial attributes
still remains challenging in real-world scenarios. For example, most
existing works \cite{rudd2016moon} predict the facial attributes
on well-cropped face images. In an extension to real-world scenarios,
those works have to utilize the face detector to localize the bounding
box of human faces before facial attribute analysis. 

Such a pipeline of face detection followed by facial attribute prediction
is nevertheless undesirable in real-world facial attribute prediction
due to two reasons. Firstly, with the preprocessing of face detector,
the capability of facial attribute prediction has to heavily relies
on the results of face detection. Unfortunately in an uncontrolled
setting \cite{liu2015deep}, the face is likely to have large out-of-plane
tilting, occlusion and illumination variations, which may affect the
algorithms of face detection and facial attribute analysis simultaneously.
For example, the Viola-Jones face detector \cite{viola2001rapid}
works well for near-frontal faces, but less effective for faces in
the pose far from frontal views. Second, face detection is heterogeneous
but subtly correlated with facial attribute prediction, and vice versa \cite{yang2015facial}.
Efficiently and effectively exploiting this correlation may help in
both detecting faces and predicting facial attributes.

We propose an end-to-end deep architecture to \emph{jointly} learning
to detect faces and analyze facial attributes. Essentially, the two
tasks share the same bottom layers (feature map in Fig. \ref{network_structure})
in the architecture to alleviate the problems of face variations aforementioned.
For instance, one face region of one image is failed
to be detected as a face due to the partial occlusion, and yet the network
can detect the existence of some facial attributes such as ``eyeglasses'',
and ``smiling''. The face region can still be used to optimize our
network and in turn help supervise the learning of the face detection
part, to improve the performance of face detection part.

In this work, the deep learning architecture is proposed for both
face detection and facial attribute analysis. The network structure
is illustrated in Fig. \ref{network_structure}. Given an entire image,
our architecture will firstly pass the image with convolutional layers
(\emph{conv1}\textendash{} \emph{conv5}) and max pooling layers to
produce the \emph{conv} feature map for face region proposals. 
The region of interest (RoI) pooling layer and face region proposal layer
are two layers introduced to facilitate the jointly learning
of our architecture. A region of interest (RoI) pooling layer pools
each face proposal on the \emph{conv5} feature map into a fixed-length
feature vector which is further processed by two fully connected layers
(\emph{fc6} and \emph{fc7}). Built on the \emph{fc7} layer, each individual
task of facial attribute analysis and face detection are jointly optimized.
By virtue of the face detection subnet, our architecture can directly
predict the facial attributes from the whole images, rather than
using the well-cropped images as previous work \cite{ranjan2016hyperface,rudd2016moon,wang2016walk}.
Extensive experiments on benchmark CelebA \cite{liu2015deep} and
LFWA \cite{huang2007labeled,liu2015deep} datasets demonstrate that our method outperforms state-of-the-art
alternatives.

\vspace{0.05in}

\noindent \textbf{Contribution}. Our main contribution is to propose
a novel deep architecture of jointly learning of face detection
and facial attribute tasks, which is capable of utilizing both tasks
to better optimize the shared network and thus improve the performance
of both tasks. Two layers \textendash{} region of interest (RoI)
pooling layer and face region proposal layer are introduced to enable learning these two tasks simultaneously. More importantly, we
can predict the facial attributes in the wild and the input images
do not need to be cropped and aligned as the standard practice in \cite{rudd2016moon,wang2016walk}.
Finally, our joint deep architecture achieves state-of-the-art result
on the biggest benchmark dataset for facial attribute analysis\textendash{}
CelebA dataset \cite{liu2015deep} and LFWA \cite{liu2015deep,huang2007labeled}
datasets. 

\section{Related Work}

\begin{figure*}[t]
\begin{centering}
 \includegraphics[scale=0.42]{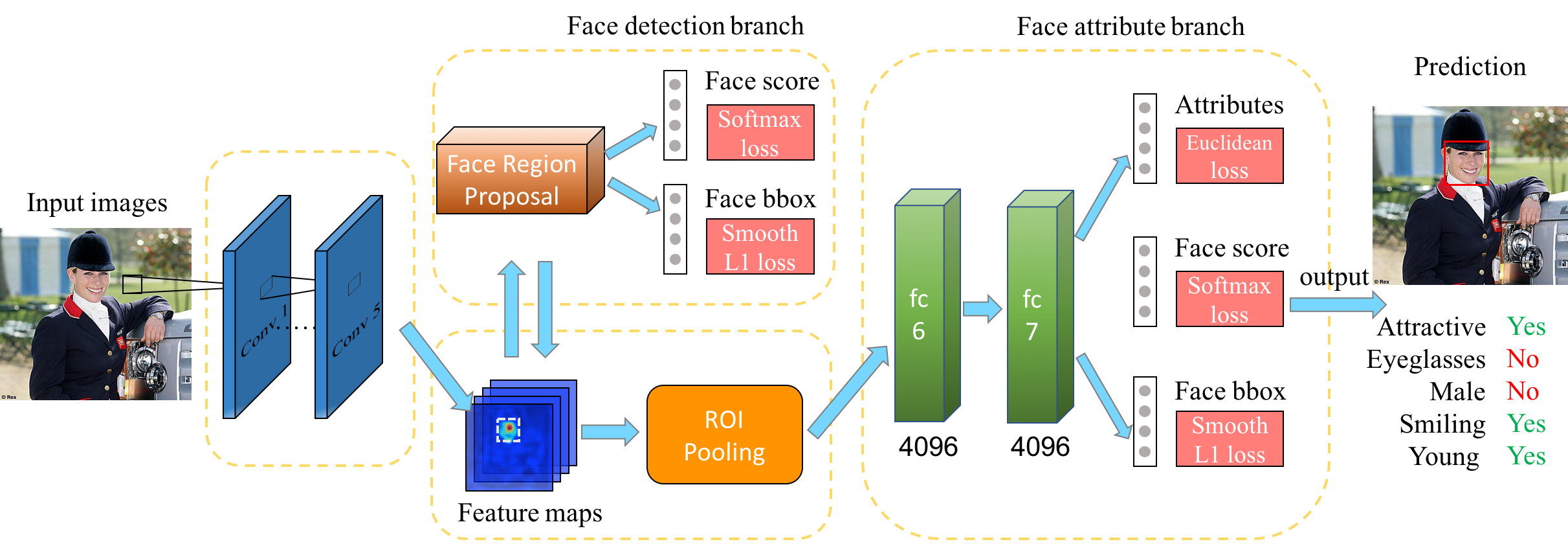}
\par\end{centering}
\caption{Overview of the proposed architecture. }
\label{network_structure} 
\end{figure*}

\noindent \textbf{Facial attribute analysis. }It was first studied
by Kumar et al. \cite{kumar2009attribute}. In term of different visual
features and distinctive learning paradigm, the facial attribute analysis
has been developed into three categories: (1) the methods \cite{kumar2009attribute}
of using hand-crafted visual features, such as SIFT \cite{lowe2004distinctive}
and LBP \cite{lbp_tpami}; (2) the methods of utilizing the recent
deep features \cite{liu2015deep,off_shelf_face,wang2016walk}; and
(3) multi-task methods of learning facial attribute \cite{abdulnabi2015multi,rudd2016moon}.
We here highlight the differences between our architecture and these
previous works. Liu \emph{et al.} \cite{liu2015deep} cascaded three
deep networks pre-trained for facial attribute prediction. In contrast,
we show that the tasks of face detection and facial attribute prediction
are highly correlated and our jointly learning architecture can improve
both tasks. Rudd \emph{et al.} \cite{rudd2016moon} introduced a mixed
objective optimization network which utilizes distribution
of attribute labels to learn each task.
Abdulnabi \emph{et al.} \cite{abdulnabi2015multi} proposed a multi-task
CNN model sharing of visual knowledge between tasks for facial attribute
analysis. Comparing with \cite{rudd2016moon,abdulnabi2015multi},
we focus on jointly learning the face detection and facial attribute
analysis; and our model can predict facial attributes on the images
in the wild \cite{rudd2016moon,wang2016walk}\emph{ i.e.} without
cropped and aligned, as illustrated in Fig. \ref{network_structure}
and Fig. \ref{Detection_result_of_joint}. 

\vspace{0.05in}

\noindent \textbf{Face detection.} The work of Viola and Jones \cite{viola2001rapid,viola2005multiple}
made face detection usable in many real world applications. The cascaded
classifiers were built on Harr-like features to detect human faces.
The deformable part model (DPM) \cite{felzenszwalb2010object} on
top of HOG feature is a general object detector and can also be used
for face detection \cite{dalal2005histograms}. Recent advances of
deep learning architectures also inspire another category of methods
for face detection. Yang \cite{yang_face} used the fully convolutional
networks (FCN) to generate the heat map of facial parts for producing
face proposals. In contrast to these works, our face detection task
gets benefit from not only the well-designed architecture (in Fig.
\ref{network_structure}), but also the jointly learning process with
facial attribute prediction. Further inspired by Fast-RCNN \cite{fast_rcnn,faster_rcnn},
we take the face detection as a special case of the general semi-rigid
object detection. More specifically, given an image, our face detection
will try to answer two questions: (1) whether this patch contains
faces or not? \emph{i.e}. face score task in Fig. \ref{network_structure}.
(2) can we detect the bounding box of faces if there is any face?
\emph{i.e}. face bounding box task in Fig. \ref{network_structure}.
The jointly optimizing these two sub-tasks will better solve face
detection.

\vspace{0.05in}

\noindent \textbf{Multi-task learning.} Our framework can be categorized
as multi-task learning \cite{yosinski2014transferable,pan2010survey},
which shares the information and explores the similarity of related
tasks on the same data. The multi-task learning can facilitate a wide
range of tasks and applications, including but not limited to action
recognition \cite{zhou2013learning}, information retrieval \cite{liu2015representation},
facial landmark detection~\cite{zhang2014facial,ranjan2016hyperface},
and facial attribute prediction ~\cite{moon_attrb,ehrlich2016facial,abdulnabi2015multi,all_in_one,lu2016fully}.
The recent technical report \cite{ranjan2016hyperface} also combine
face detection with the tasks of locating face landmarks and recognizing
gender. However, unlike our facial attribute prediction on the images
in the wild, their work still has to firstly detect face regions for
the facial landmarks and predicting gender. 

\section{Our Deep Architecture }

In this section, we firstly overview our network in Sec. \ref{subsec:Overview},
and then the tasks are defined in Sec. \ref{subsec:Task-formulation}.
The Face region proposal and RoI pooling layers are explained accordingly
in Sec. \ref{subsec:Face-Region-proposal} and Sec. \ref{subsec:The-RoI-pooling}.
Finally, we utilize the network to solve the facial attribute prediction
and face detection in Sec. \ref{subsec:Facial-attribute-analysis}. 

\subsection{Overview\label{subsec:Overview}}

Figure \ref{network_structure} shows our framework for jointly face
detection and attribute analysis. Our architecture takes an entire
image as input and a set of face bounding boxes as labels for training. The whole
network firstly processes the image with several convolutional layers
(\emph{conv1}\textendash{} \emph{conv5}), and max pooling layers to
produce a \emph{conv} feature map for face region proposal. For feature
map of each proposed face, we employ a region of interest (RoI) pooling
layer to pool it into a fixed-length feature vector. Each feature
vector is further processed by two fully connected layers (\emph{fc6}
and \emph{fc7}) and thus used for the tasks of facial attribute analysis
and face detection. On top layers, our architecture has \emph{face
detection branch} and \emph{facial attribute branch}. 

Our network follows the art and design of VGG-16 \cite{returnDevil2014BMVC}.
Particularly, the kernel size, stride and the number of filters in
convolutional layers (\emph{conv1}\textendash{} \emph{conv5}) and
the two fully connected layers (\emph{fc6} and \emph{fc7}) are exactly
the same as the corresponding layers in VGG-16 architecture.

\subsection{Task formulation\label{subsec:Task-formulation}}

Suppose we have the labelled source training dataset $\mathcal{D}_{s}=\left\{ \mathbf{I},\mathbf{a},\mathbf{L}\right\} $
with $N$ training instances and $M$ attributes. $\mathbf{I}$ denotes
the patches of training images and $\mathbf{L}$ denotes the labels. We use the $\mathbf{L}$ matrix to
both denote whether an image patch contains the face, and whether
a facial attribute exists in the image patch. Particularly, for the
$i$-th image patch $\mathbf{I}_{i}\:(i=1,\cdots,N)$, we use $\mathbf{L}_{i\star}=\mathbf{0}$
to indicate that this image patch does not contain a human face. If
$\mathbf{I}_{i}$ is a face image patch, we use $\mathbf{L}_{ij}=+1$
to denote the existence of $j-$th facial attribute $\mathbf{a}_{j}\,(j=1,\cdots,M)$;
$\mathbf{L}_{ij}=-1$ otherwise. 

As illustrated in Fig. \ref{network_structure}, our network extracts
the image patch to a $4096-dim$ feature vector, and it is denoted
as $f\left(\mathbf{I};\Theta\right)$, where $\Theta$ is the parameter
set of the deep architecture. We have the prediction tasks of facial
attribute, face score and face bounding box. Each task has their own
parameters on the last layer.

\vspace{0.05in}

\noindent \textbf{Facial attribute prediction.} To predict the attributes
of the image $\mathbf{I}^{*}$, we need to learn a function $\mathbf{L}_{i\mathbf{a}}^{*}=\Psi\left(\mathbf{I}^{*}\right)$
to predict the facial attribute $\mathbf{a}$. We thus have the predicting
function $\Psi=\left[\psi_{i}\right]_{i=1,\cdots,M}$, and $\psi_{i}\left(\mathbf{I}\right):\mathbb{R}^{4096}\rightarrow\left\{ +1,-1\right\} $;
specifically, we consider $\Psi\left(x\right)=W_{a}^{T}x$, where
$W_{a}\subseteq\mathbb{R}^{4096\times40}$ for all $40$ attributes.

\vspace{0.05in}

\noindent \textbf{Face score}. This task aims to predict the score
whether an image patch is face. Essentially, we need to learn $\Phi\left(x\right):\mathbb{R}^{4096}\rightarrow\left\{ +1,-1\right\} $,
and $\Phi\left(x\right)=W_{s}^{T}x$, where $W_{a}\subseteq\mathbb{R}^{4096\times2}$
for binary prediction task.

\vspace{0.05in}

\noindent \textbf{Face bounding box}. We regress the bounding box
of faces by $\Omega\left(x\right):\mathbb{R}^{4096}\rightarrow\mathbb{R}^{4}$,
and we configure the form as $\Omega\left(x\right)=W_{b}^{T}x$, and
$W_{b}\subseteq\mathbb{R}^{4096\times4}$ for the new bounding box.

\subsection{Face region proposal layer \label{subsec:Face-Region-proposal}}

Inspired by the work of object detection \cite{faster_rcnn}, our
network has the branch of face region proposal. Specifically, this
branch takes an image of any size as input and outputs a set of rectangular
face proposals. To generate the facial region proposal, we slide the
branch over the convolutional feature map after the five convolutional
layers (\emph{conv1}\textendash{} \emph{conv5}) in Fig. \ref{network_structure}.
The extracted \emph{conv5} features are further fed into two sibling
fully connected layers, i.e. \emph{face bounding box} (bbox) and \emph{face
score} layer. Face bounding box layer employs the smooth $L_{1}$
loss (defined in \cite{fast_rcnn}) enables a regressor to predict
the facial bounding box. Face score layer utilizes the softmax loss
to indicate whether the bounding box is a face. For each image, we
selected top-300 face region proposals in term of the face scores
computed.

\subsection{The RoI pooling layer\label{subsec:The-RoI-pooling}}
The RoI pooling layer can convert the feature maps of face region
proposal (with the size of $h\times w$) to a fixed spatial extent
of $H\times W$, which facilitates the further process. Here, $h,w,H,W$
are heights and widths of each rectangle region of feature map in
Fig. \ref{network_structure}. For the varying size of input feature patches,
we vary the size of filters of pooling layer with the sub-window of
approximate size $h/H\times w/W$. The RoI pooling technique is 
 adopted from the work of object detection \cite{fast_rcnn} which inspired by the SPPnets \cite{he2014spatial}.

\subsection{Facial attribute analysis and face detection \label{subsec:Facial-attribute-analysis}}

The fixed-length feature vector extracted from RoI Pooling layer is
further fed into two fully connected layers (\emph{fc6} and \emph{fc7})
in order to enable the tasks of facial attribute analysis. Particularly,
built on the layers of \emph{fc6} and \emph{fc7, }the multi-task network
structure is employed to analyze each attribute individually. This
can be modeled as the minimization of the expected loss over all the
training instances which is

\begin{align}
\left\{ \Theta,W_{a},W_{s},W_{b}\right\}  & =\underset{}{\mathrm{argmin}}\:\mathcal{L}\left(\mathbf{I};\Theta,W_{a},W_{s},W_{b}\right)\label{eq:baseline_model}
\end{align}

\noindent where $\mathcal{L}\left(\mathbf{I};\Theta,W_{a},W_{s},W_{b}\right)$
is the loss function of jointly learning; and we have 
\begin{align}
\mathcal{L}\left(\mathbf{I};\Theta,W_{a},W_{s},W_{b}\right) & =\lambda_{1}\mathcal{L}_{a}\left(\Phi\left(f\left(\mathbf{I}\right)\right)\right)+\lambda_{2}\mathcal{L}_{s}\left(\Psi\left(f\left(\mathbf{I}\right)\right)\right)\label{eq:loss_func_split}\\
 & +\lambda_{3}\mathcal{L}_{b}\left(\Omega\left(f\left(\mathbf{I}\right)\right)\right)\nonumber 
\end{align}
here the loss function $\mathcal{L}_{a}\left(\Phi\left(f\left(\mathbf{I}\right)\right)\right)$
on \emph{facial attribute prediction} is the mean square error loss
of all attributes; we use softmax loss $\mathcal{L}_{s}\left(\Psi\left(f\left(\mathbf{I}\right)\right)\right)$
for \emph{face score task,} and finally the smooth $L_{1}-$ loss
$\mathcal{L}_{b}\left(\Omega\left(f\left(\mathbf{I}\right)\right)\right)$
\cite{fast_rcnn} is employed to regress the \emph{face bounding box}.

\noindent We denote the shared parameters of deep architectures as
$\Theta$ which are jointly optimized by all these three tasks. Since
these three tasks are highly related, learning in such a way can not
only greatly reduce the prediction error of each individual task,
but also accelerate the convergence rate of learning the whole network.

For the testing image $\mathbf{I}_{k}$, the predicted result $\hat{\mathbf{L}}_{kj}$
of the $j-$th facial attribute $\mathbf{a}_{j}\,(j=1,\cdots,M)$
is thresholded by

\begin{equation}
\hat{\mathbf{L}}_{kj}=\begin{cases}
1 & \psi\left(\mathbf{I}_{k}\right)>\tau\\
-1 & \psi\left(\mathbf{I}_{k}\right)\leq\tau
\end{cases}\label{eq:threshold}
\end{equation}

\noindent where $\tau$ is the threshold parameter.

\begin{figure*}
\begin{centering}
\includegraphics[scale=0.38]{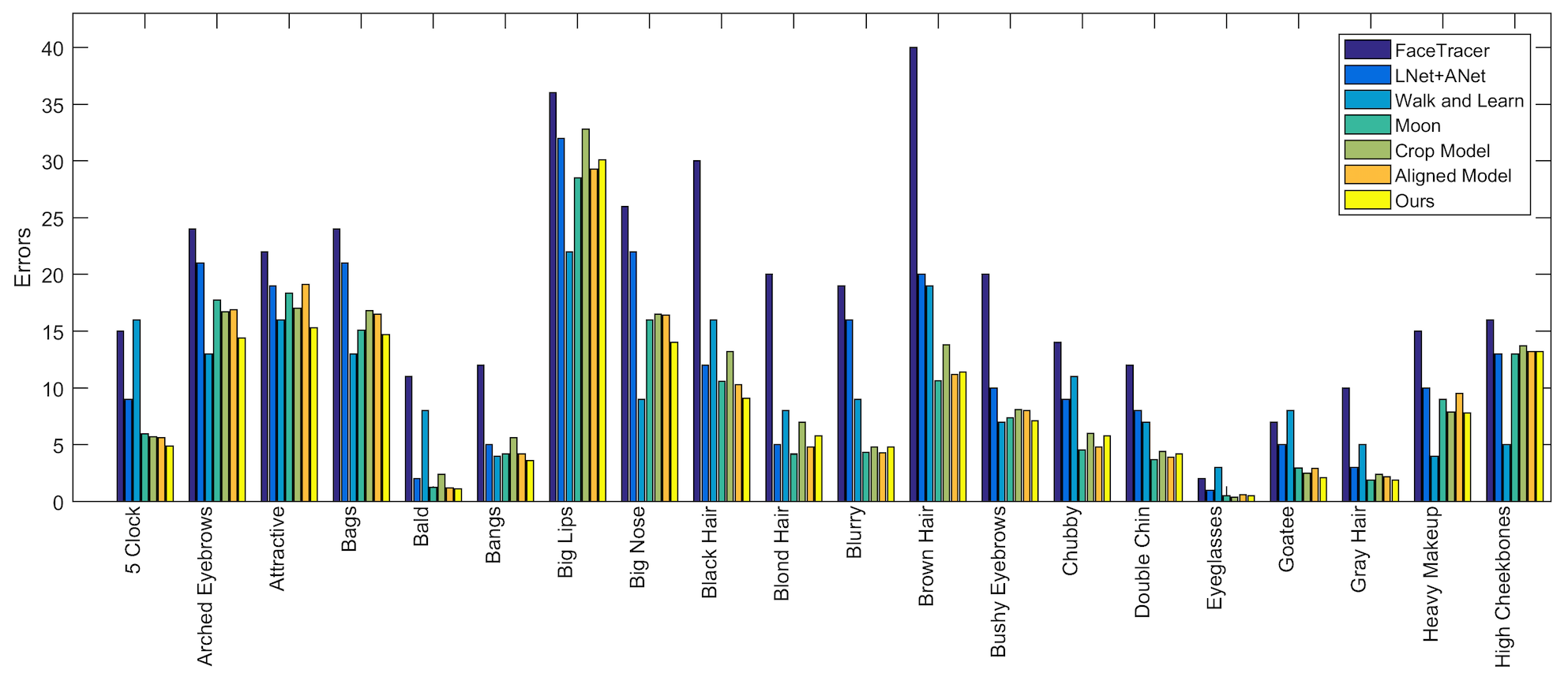} 
\par\end{centering}
\begin{centering}
\includegraphics[scale=0.38]{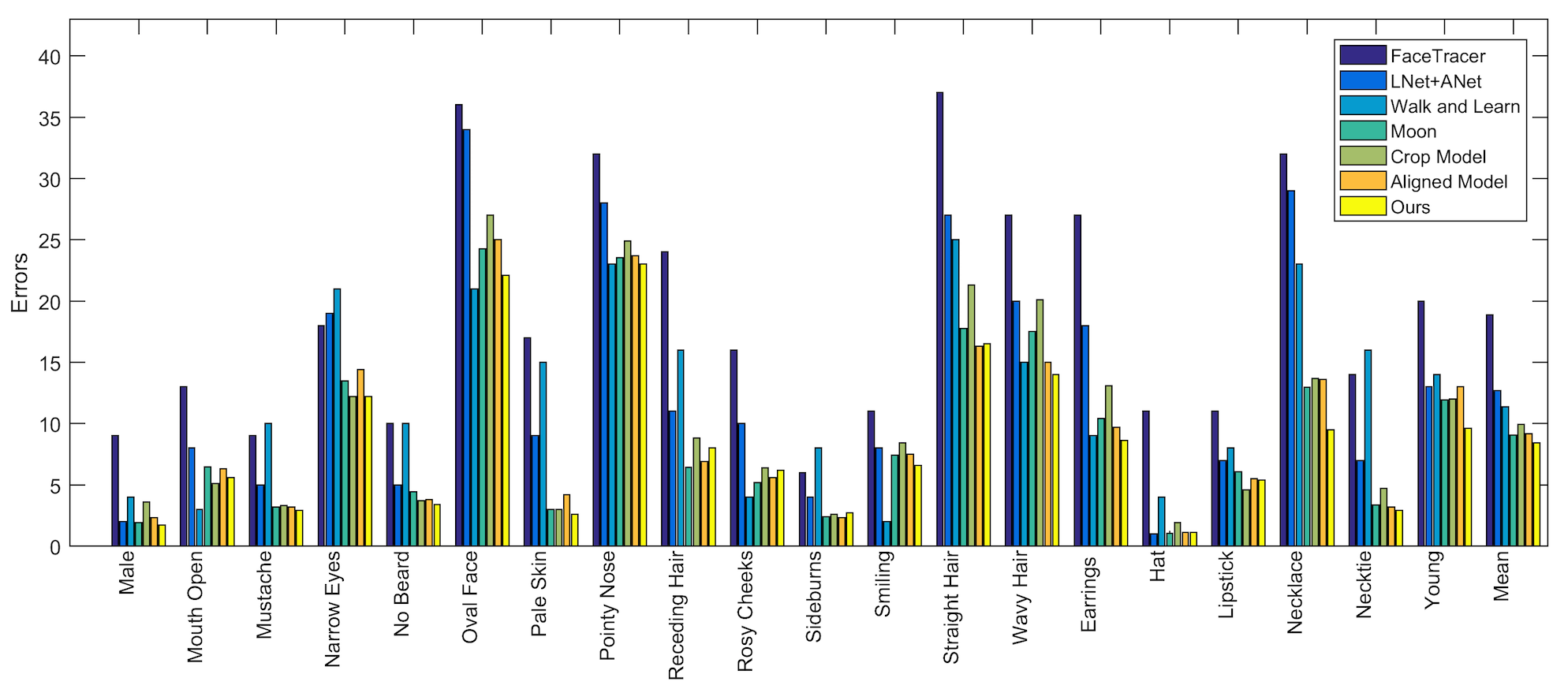} 
\par\end{centering}
\caption{Performance comparison with state-of-the-art methods on CelebA on
all 40 attributes. X-axis indicates each attribute, while y-axis is
the error rate of attribute prediction (the lower value, the better
performance of one method). The averaged error rate of FaceTracer,
LNets+ANet, Walk and Learn, Moon, Baseline (Cropped images), Baseline
(Aligned images), and ours are $18.88\%$, $12.70\%$, $11.35\%$,
$9.06\%$, $9.95\%$, $9.19\%$ and $\textbf{8.41\%}$ respectively.
\label{fig:celebA-Performance-comparison-with}}
\end{figure*}

\begin{figure*}
\begin{centering}
\includegraphics[scale=0.38]{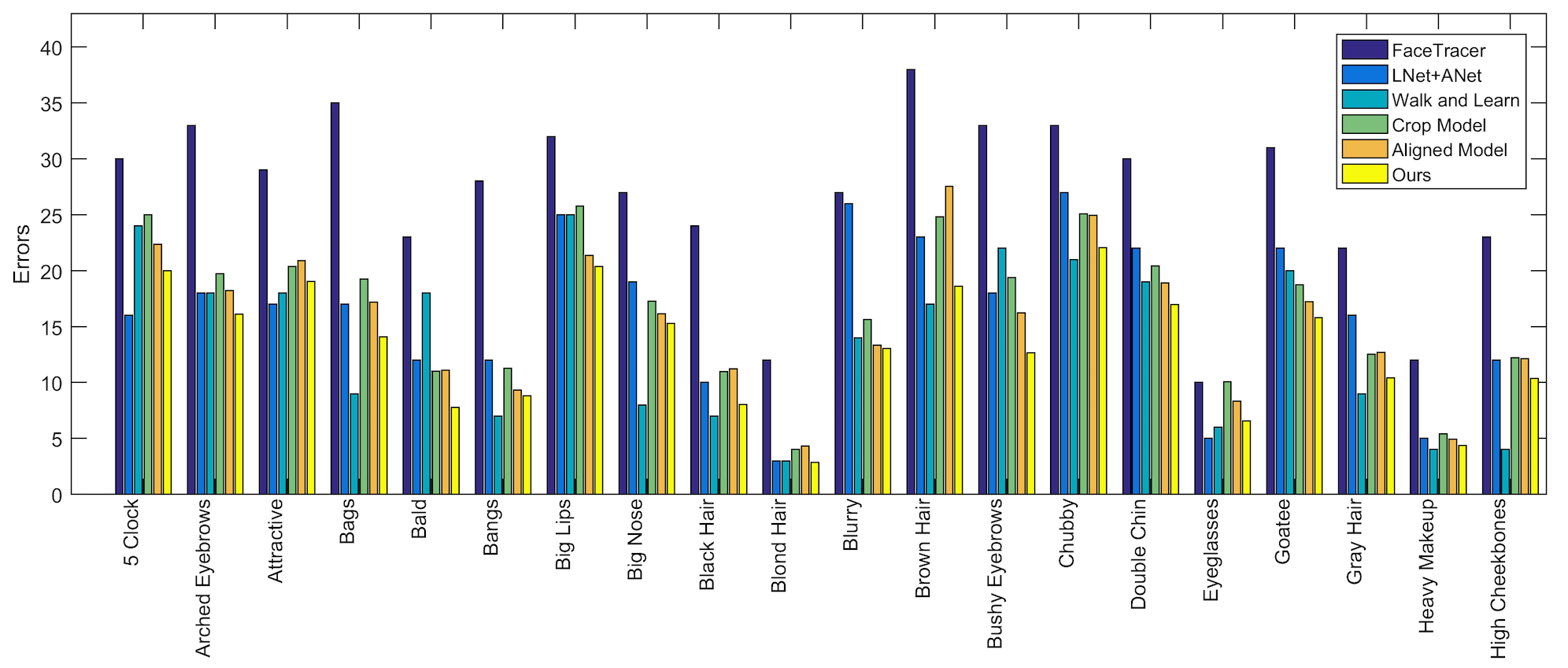} 
\par\end{centering}
\begin{centering}
\includegraphics[scale=0.38]{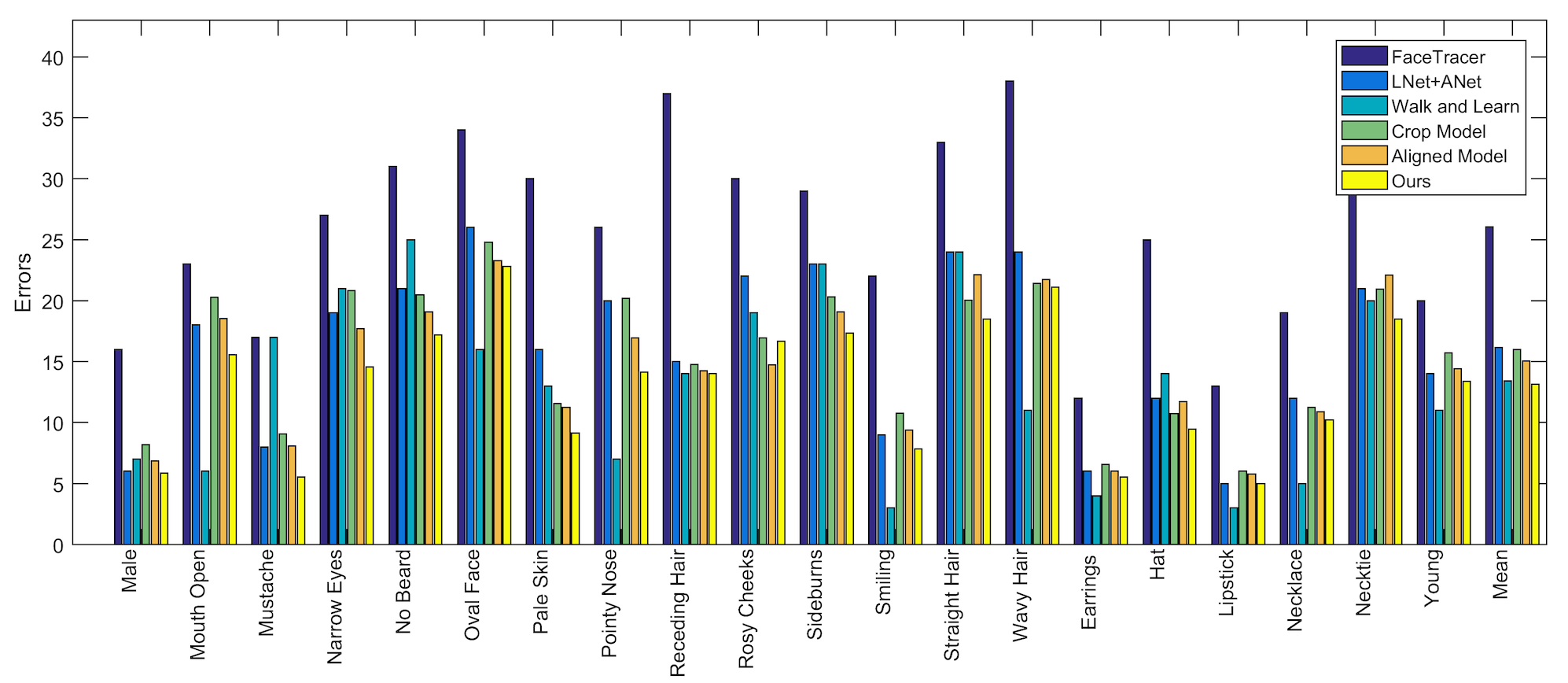} 
\par\end{centering}
\caption{Performance comparison with state-of-the-art methods on LFWA on all
40 attributes. The averaged error rate of FaceTracer, LNets+ANet, Walk
and Learn, Baseline (Cropped images), Baseline (Aligned images), and
ours are $26.07\%$, $16.15\%$, $13.40\%$, $15.99\%$, $15.06\%$
and  $\textbf{13.13\%}$ respectively. \label{fig:Performance-comparison-lfwa}}
\end{figure*}

\section{Experiments}

\label{Experiments}

\subsection{Datasets and settings}

\begin{figure}[t]
\begin{centering}
\includegraphics[width=70mm]{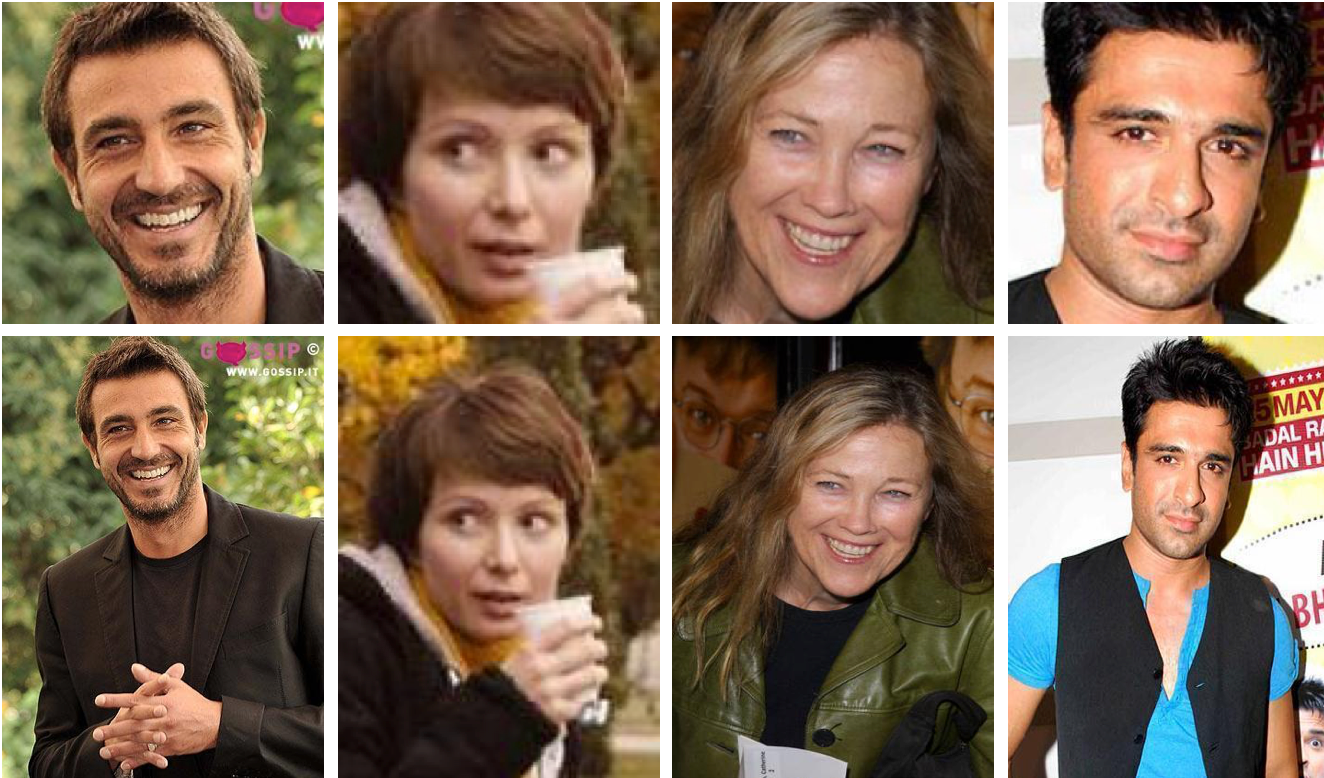} 
\par\end{centering}
\caption{\label{fig:train-images} The first row is aligned images provided by the CelebA. 
The second row is CelebA raw images. We train our end-to-end
model on raw images.}
\end{figure}

\noindent We conduct the experiments on the CelebA dataset \cite{liu2015deep}
and LFWA dataset \cite{huang2007labeled,liu2015deep}.

\noindent \textbf{CelebA \cite{liu2015deep}} contains approximately
200k images of 10k identities. Each image is annotated with $5$ landmarks
(two eyes, the nose tips, the mouth corners) and binary labels of
$40$ attributes. To make a fair comparison with the other facial
attribute methods, the standard split is used here: the first $160k$
images are used for training, $20k$ images for validation and remaining
$20k$ for testing. CelebA provides two types of training images,
\emph{i.e.}, aligned and cropped face images and raw images as shown
in the first and second row of Fig. \ref{fig:train-images} respectively.
We use the raw images for training our joint deep architecture. Since
there is no ground-truth bounding box label for faces in raw images,
the face detector \cite{dalal2005histograms,dlib09} is employed here
to help generate the face bounding box for training images. We use
the implementation of dlib toolbox \cite{dlib09,dalal2005histograms}
for generating labels. Note that face bounding box generation is only required in the training stages 
for synthesizing labels.
At testing phrase, raw images are directly
input for both facial attribute analysis and face detection.

\vspace{0.05in}

\noindent \textbf{LFWA \cite{liu2015deep}} is constructed based on
face datasets LFW \cite{huang2007labeled}. It contains approximately
$13143$ images of $10k$ identities. Following \cite{liu2015deep},
$50\%$ of the images for training, and the other $50\%$ are used
for testing. LFWA has $40$ binary facial attributes, the same as
CelebA. We also generate face bounding box for LFWA to train our joint
learning network.

\vspace{0.05in}

\noindent \textbf{Evaluation metrics.} We take the attribute prediction
as classification tasks and thus mean accuracy can be computed. Particularly,
we evaluate the performance as comparable to \cite{rudd2016moon,liu2015deep}
by the mean error which is defined as $mean\:error=1-mean\,accuracy$.

\vspace{0.05in}

\noindent \textbf{Implementation and Parameter settings.} The $\tau$
is set as $0$ in Eq (\ref{eq:threshold}); and square error loss
is used in Eq (\ref{eq:baseline_model}). We empirically set $\lambda_{1}=\lambda_{2}=1,\lambda_{3}=2$
in Eq (\ref{eq:loss_func_split}). The convolutional layers and fully
connected layers are initialized by the 16 layer VGG network \cite{returnDevil2014BMVC}
individually. We use the open source deep learning framework Caffe
\cite{caffe} to implement our structure. A single end-to-end model
is used for all the testing. We employ the stochastic gradient descent
to train our network. Dropout is used for fully connected layers and
the ratio is set to $0.5$. For training CelebA dataset, with initial
learning rate $0.001$, and gradually decreased by $1/10$ at $100k$,
$150k$ iterations, the total training iterations are $180k$. For
training LFWA dataset, as there are only limited training images -
$6263$, We fine-tune from pre-trained CelebA model, with initial
learning rate $0.0001$, and decreased by $1/10$ at $40k$ iterations,
the total training iterations is $60k$. 
Once trained, our framework
can predict the facial attributes and detect faces on the images in
the wild for any testing images. 

\vspace{0.05in} 

\noindent \textbf{Running cost.} Our facial model get converged with
$180k$ iterations and it takes $29$ hours on CelebA
with one NVIDIA TITANX GPU. 
On LFWA dataset, Our facial model get converged with $60k$
iterations and takes $9$ hours. For training
all the model, and it takes around $4$
GB GPU memory.

\subsection{Competitors. }

Our model is compared against state-of-the-art methods and several
baselines. 
Particularly, (1) \textbf{FaceTracer} \cite{kumar2008facetracer}
is one of the best methods with the hand-crafted features. The features
used including HOG \cite{PHOG2007CVIR} and color histograms of facial
regions of interested to train SVM classifier for predicting facial
attributes. (2)\textbf{ LNets+ANet} \cite{liu2015deep} migrates two
deep CNN face localization networks to one deep CNN network for facial
attribute classification. (3) \textbf{Walk and Learn} \cite{wang2016walk}
learns good representations for facial attributes by exploiting videos
and contextual data (geo-location and weather) as the person walks.
(4)\textbf{ Moon} \cite{rudd2016moon} is a mixed objective optimization
multi-task network to learn all facial attributes, achieves the best result
on CelebA dataset. (5) \textbf{Cropped model }is a variant of our
model without using the face detection branch. The raw image is also
used as the input and we crop the faces from the images by Dlib toolkit
to train facial attribute models. The processed face images are utilized
to train the model. (6)\textbf{ Aligned model} uses the aligned and cropped images provided
by CelebA and LFWA to train facial attribute model, which is used
by most of the state-of-the-art methods. To make a fair comparison, these
two baseline models are initialized by 16 layer VGG network; same
as the proposed structure.

\begin{table}
\centering{}%
\begin{tabular}{c|c|c}
\hline 
Methods  & CelebA (\%)  & LFWA (\%) \tabularnewline
\hline 
\hline 
FaceTracer \cite{kumar2008facetracer} &18.88 &26.07 \tabularnewline
\hline 
LNets+ANet \cite{liu2015deep}  &12.70 &16.15  \tabularnewline
\hline 
Walk-and-Learn \cite{wang2016walk} &11.35  &13.40 \tabularnewline
\hline 
Moon \cite{moon_attrb}  &9.06  & - \tabularnewline
\hline 
Baseline: Cropped Model  &9.95 &15.99   \tabularnewline
Baseline: Aligned Model   &9.19 &15.06  \tabularnewline
\hline
Ours &\textbf{8.41} & \textbf{13.13}   \tabularnewline
\hline 
\end{tabular}\caption{ \label{tab:Comparison-of-mean-error} Comparison of mean error
 on CelebA and LFWA datasets.}
\end{table}

\begin{figure*}[tp]
\begin{centering}
\includegraphics[scale=0.24]{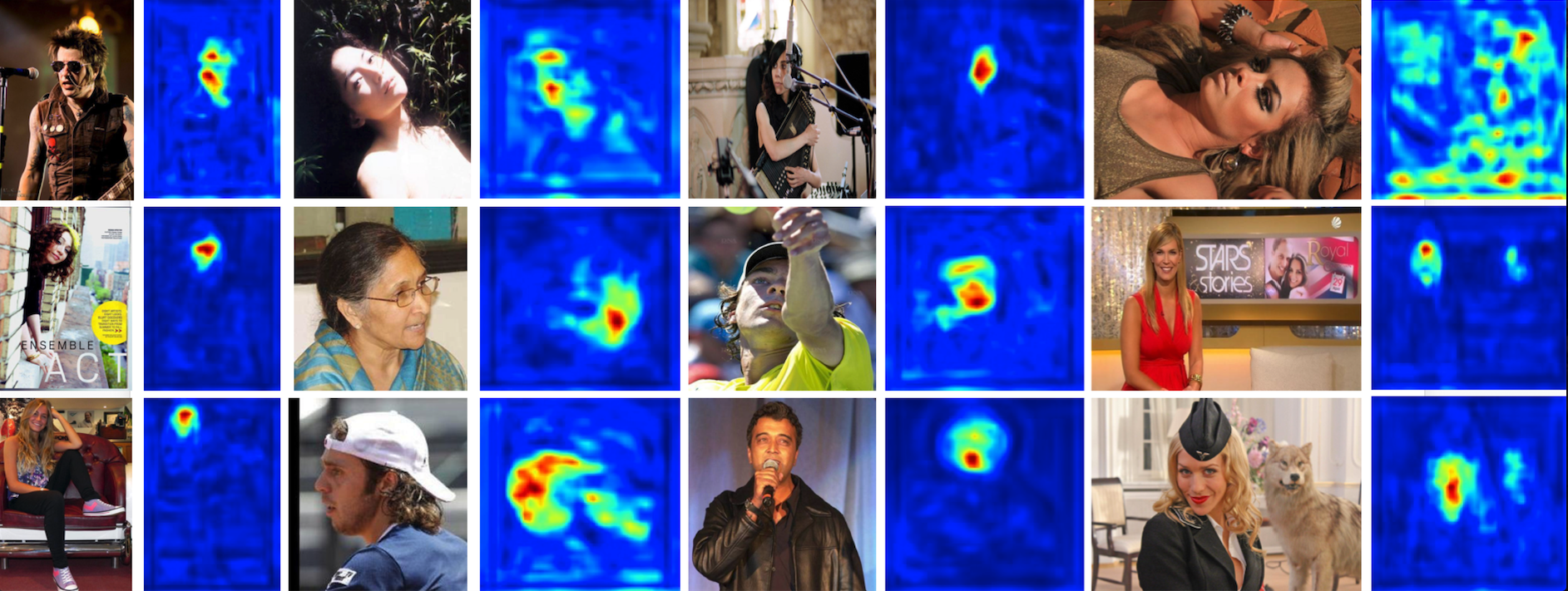} 
\par\end{centering}
\caption{\label{Detection_result_of_joint} The visualization of feature map
for \emph{conv5} layer. The images are from CelebA test set. The different
types of images are shown in the first three columns. These images
have large pose variation of heads and high occlusion. Some failed
examples are shown in the last column.}
\end{figure*}

\subsection{Comparison with the State of the Art}
We compared our model with state-of-the-art methods: FaceTracer \cite{kumar2008facetracer},
LNets+ANet \cite{liu2015deep}, Walk and Learn \cite{wang2016walk},
Moon \cite{rudd2016moon} and two baselines. The results on the testing
split of CelebA and LFWA are reported in Tab. \ref{tab:Comparison-of-mean-error}, Fig. \ref{fig:celebA-Performance-comparison-with}
and Fig. \ref{fig:Performance-comparison-lfwa}, respectively. 
Note that please refer to the supplementary material for the full comparison results on each attribute. Comparing
with all the competitors, we draw the following conclusions.

\vspace{0.05in}

\noindent \textbf{Our model beats all the other methods by the mean error
on CelebA and LFWA dataset. }As we can see from Fig. \ref{fig:celebA-Performance-comparison-with},
our approach obtains the mean error $\textbf{8.41\%}$ which is the
lowest among all the competitors, and it outperforms all the other
methods on CelebA dataset. And on LFWA dataset as shown in Fig. \ref{fig:Performance-comparison-lfwa},
our mean error is only $\textbf{13.13\%}$ which is the lowest among
all the competitors. This validates the efficacy of our joint learning
architecture. Particularly, We compare the classification error
in each individual attribute in Fig. \ref{fig:celebA-Performance-comparison-with}
and Fig. \ref{fig:Performance-comparison-lfwa}. We note that on more
than half among the total 40 attributes, our framework is significantly
better than the other competitors, since our jointly learning architecture
can efficiently leverage the information between face detection and
facial attribute prediction. Particularly, comparing with the other
works, all our attribute prediction tasks share the same deep learning
architecture\emph{, i.e. conv1}-\emph{conv5}, feature map and \emph{fc6},
\emph{fc7} layers as illustrated in Fig. \ref{network_structure}).
These shared structures implicitly model the correlations between
each attribute task. Furthermore, on gender attribute
prediction, our framework can achieve $1.7\%$ error on CelebA; in contrast,
the error of HyperFace \cite{ranjan2016hyperface} is $3.0\%$.
That indicates that our results can get $1.3\%$ improvement over
that of HyperFace \cite{ranjan2016hyperface}.

\vspace{0.05in}

\noindent \textbf{Our model can process the images in the wild. }Our face region proposal and RoI pooling layers are flexible
enough to directly process the images in the wild. This thus better
demonstrates the effectiveness of our models. Specifically, our model
not only achieves the best performance on the benchmark dataset \textendash{}
CelebA and LFWA, but also our tasks of facial attribute prediction
do not need to align and crop the facial images as have done in many
previous work \cite{rudd2016moon,kumar2008facetracer,liu2015deep}.
Additionally, unlike the work of Walk and Learn \cite{wang2016walk},
our model is not trained by the external data; and yet still obtains
better results than Walk and Learn \cite{wang2016walk}.

\vspace{0.05in}

\noindent \textbf{Our model is very efficient in term of jointly
learning to detect faces and predict facial attributes. }Our joint
learning results have greatly improved over those of two baseline
models \textendash{} Cropped, Aligned models. These two methods
are yet another two naive baselines of directly learning the facial
attribute tasks. Particularly, on CelebA dataset, our model beats
the two methods on the classification error of $27$ attributes
(totally $40$ facial attributes) as compared in Fig. \ref{fig:celebA-Performance-comparison-with}.
Figure \ref{fig:Performance-comparison-lfwa} shows that our joint
model hit better results than the two baseline models on all the $40$
attributes. This reveals attribute prediction tasks get benefit from
the face detection task. Also note that unlike LNets+ANet \cite{liu2015deep}
using two networks to localize the face and another one network to
extract features, we can train our model by a single end-to-end network
here since the shared architectures (\emph{conv1 \textendash{} conv5})
can explicitly model the important facial parts, again thanks to the
face detection subnet. 

\vspace{0.05in}

\noindent \textbf{Qualitative results. }The important facial parts
extracted by the shared feature map are visualized in Fig. \ref{Detection_result_of_joint}.
In particular, four groups of images are shown; and the feature map
has higher activation on the regions of human faces. The first 3 columns
are success cases, even thought the images have large pose (Column
2) or very high occlusion (Column 3). We also list some failure examples
in Column 4, which are some extreme cases. Some of them are caused
by ambiguous or too small view (e.g. the cat face is also similar
to human's). The feature map reveals our model pay attention on the
important facial parts which helps to analyze facial attributes.

\subsection{Results of Face Detection}

Our face detection is compared against the face detector of dlib toolbox
\cite{dlib09}, which is an implementation of the generic object detector
\cite{dalal2005histograms} on face images. On CelebA dataset, there
are only 5 landmarks of faces, but no labeled ground-truth bounding
box of faces for raw images. We generate the ground truth bounding box by 5 landmark coordinates for the test set. If the IOU (Intersection over Union) of predicted bounding box and the ground truth bounding box is larger than 0.5, we assume the prediction is correct.
We compare the two detectors on the testing split of CelebA.

\begin{figure}
\begin{centering}
\includegraphics[width=80mm]{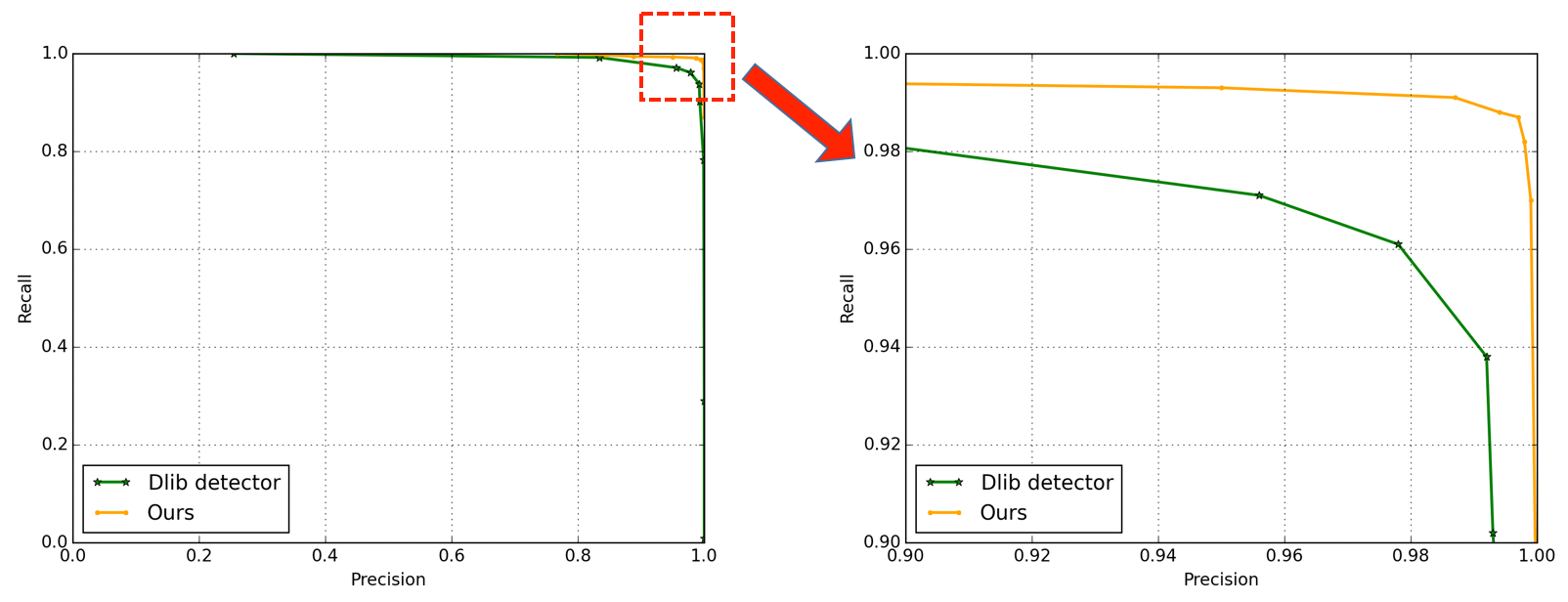} 
\par\end{centering}
\caption{\label{Comparison_of_detector}Precision-Recall (PR) results of face
detection. }
\end{figure}

The results are shown in Fig. \ref{Comparison_of_detector}. We compare
the Precision-Recall curve for two methods. We find that both the
Dlib detector and our face detector can achieve very high face detection
on the CelebA test set as shown in the left subfigure of Fig. \ref{Comparison_of_detector},
which shows the efficacy of both detectors of solving the task of
face detection. Nevertheless, our face detector still beats the Dlib
detector by a relatively large margin. The AUC values of Dlib detector
and ours are $0.938$ and $0.982$ respectively. To better show the
difference, we highlight the up-right corner of PR curve in the right
subfigure as shown in Fig. \ref{Comparison_of_detector}. This result
further proves that our jointly learned architecture can make the
face detection and face attribute tasks help each other via sharing
the same parameters of the deep network. This shows that attribute learning
does help for face detection also it provides more detailed information
about the face.

\section{Conclusion}

In this paper, we propose a novel joint deep architecture for facial
attribute prediction and face detection. Different from the previous pipeline
of face detection followed by facial attribute prediction, our architecture
takes an entire image as input, enables both face detection and facial
attribute analysis. The proposed architecture can not only exploit
the correlation of face detection and face attribute prediction, but
also boost both tasks. The experimental results on CelebA and LFWA datasets
show the efficacy of proposed methods over the other state-of-the-art
methods.

{\small
\bibliographystyle{ieee}
\bibliography{egpaper_final}
}

\end{document}